\documentclass{bmvc2k}

\title{\Large{\: Unconditional Image-Text Pair Generation with \\ \: Multimodal Cross Quantizer}}

\addauthor{\quad Hyungyung Lee}{ttumyche@kaist.ac.kr}{1}
\addauthor{\quad Sungjin Park}{zxznm@kaist.ac.kr}{1}
\addauthor{\quad Joonseok Lee}{joonseok2010@gmail.com}{2, 3}
\addauthor{\quad Edward Choi}{edwardchoi@kaist.ac.kr}{1}

\addinstitution{
 KAIST
}
\addinstitution{
 Google Research
}

\addinstitution{
 Seoul National University
}

\runninghead{LEE ET AL.}{UNCONDITIONAL IMAGE-TEXT PAIR GENERATION WITH MXQ}

\usepackage{xspace}
\usepackage{mathtools}
\usepackage{enumitem}
\usepackage{booktabs}
\usepackage{multirow}
\usepackage{adjustbox}
\usepackage{nccmath}
\usepackage{bm,amsmath}
\usepackage{xspace}
\usepackage{graphicx}
\usepackage{color}

\usepackage{colortbl}
\definecolor{light_gray}{RGB}{170,170,170}

\usepackage[utf8]{inputenc} 
\usepackage[T1]{fontenc}    
\usepackage{url}            
\usepackage{booktabs}       
\usepackage{amsfonts}       
\usepackage{nicefrac}       
\usepackage{microtype}      
\usepackage{longtable,lscape}
\usepackage{array,multirow}
\usepackage{enumitem}

\usepackage{wrapfig}

\DeclareMathOperator*{\argmin}{argmin}

\newcommand{\model}{MXQ-VAE\xspace}

\begin{document}

\maketitle

\begin{abstract}
Although deep generative models have gained a lot of attention, most of the existing works are designed for unimodal generation.
In this paper, we explore a new method for unconditional image-text pair generation.
We design Multimodal Cross-Quantization VAE (\model), a novel vector quantizer for joint image-text representations, with which
we discover that a joint image-text representation space is effective for semantically consistent image-text pair generation.
To learn a multimodal semantic correlation in a quantized space,
we combine VQ-VAE with a Transformer encoder and apply an input masking strategy.
Specifically, \model accepts a masked image-text pair as input and learns a quantized joint representation space, so that the input can be converted to a unified code sequence, then we perform unconditional image-text pair generation with the code sequence.
Extensive experiments show the correlation between the quantized joint space and the multimodal generation capability on synthetic and real-world datasets. In addition, we demonstrate the superiority of our approach in these two aspects over several baselines.
The source code is publicly available at: https://github.com/ttumyche/MXQ-VAE.
\end{abstract}

\section{Introduction}
Deep generative models focus mainly on unimodal generation, either unconditional (GAN \cite{goodfellow2014generative}, VAE \cite{kingma2013auto}, GPT \cite{brown2020language}) or conditional (VQGAN \cite{esser2021taming}, DALL-E \cite{ramesh2021zero}).
Despite these influential works, studies on multimodal generation are still uncharted. One previous work \cite{shimanuki2019joint} proposed generating image and text at the same time with a GAN-based approach.
However, the core idea was to treat the text as an image, where the model generates two images, one for the image and another for the text. Thus, this process must undergo the OCR process \cite{smith2007overview}.

In this paper, we design Multimodal Cross-Quantization VAE (\model), a novel vector quantizer that learns image-text representations to jointly generate image-text pairs without any conditional input and post-processing (\textit{e.g.}, OCR), with which we discover that a joint representation space is effective for semantically consistent image-text pair generation.

To improve a multimodal semantic correlation in a quantized space, we combine VQ-VAE \cite{oord2017neural} with a Transformer encoder \cite{vaswani2017attention} and further apply an input masking.
\model learns to discreteize masked image-text pairs into a quantized joint representation space and reconstruct them.
Specifically, the Transformer encoder learns joint representations by performing a multi-head attention across the input, thereby can capture the semantic correlation between image and text.
The input masking further enhances the correlation by making the masked part refer to the other modality to reconstruct the original input.
Thus, we can convert the input to a unified code sequence, then train Autoregressive Transformer \cite{radford2019language} to model a joint distribution over the sequence, allowing semantically consistent image-text pair generation.

We evaluate \model on one synthetic text-augmented MNIST, called Caption MNIST and three public benchmarks: Oxford Flower-102 \cite{nilsback2008automated}, CUB-200-2011 \cite{wah2011caltech}, and COCO \cite{lin2014microsoft}. We observe that \model generates semantically consistent image-text pairs better than several baselines. Specifically, our approach achieves the highest average scores of 99.2\% on Caption MNIST, outperforming the second highest baseline by $+$ 4.7\% and also improves the performance by $+$ 0.8\% on Flower, $+$ 5.3\% on CUB and $+$ 2.3\% on COCO.

In addition, to study the effectiveness of the quantized joint space for generating semantically consistent image-text pairs, we construct a corrupted dataset, called Degree dataset, by gradually adjusting the degree of alignment between image and text.
The experimental result demonstrates that our approach can uphold the semantic correlation between image and text, while baselines fail. Furthermore, we show that this result leads to semantically consistent image-text pair generation.

Contributions of this paper can be summarized as follows:
\begin{itemize}[noitemsep, nolistsep]
    \item \textbf{Unconditional Image-Text Pair Generation}: We propose for the first time a novel vector quantization method, \model, that learns the quantized joint representation space for unconditional image-text pair generation.
    \item \textbf{Semantic Consistency of the Generated Samples}: Our experimental results reveal that \model generates a semantically consistent image-text pairs on multiple benchmark datasets, including Caption MNIST, Oxford Flower-102, CUB-200-2011, and COCO against several baselines.
    \item \textbf{Multimodal Semantic Correlation}: Additionally, the experimental results on the Degree dataset demonstrate that \model learns the meaningful semantic correlation between image and text in the quantized joint space.
    Furthermore, it turns out that the quantized joint space leads to semantically consistent image-text pair generation.
\end{itemize}

\section{Related works}
\noindent
\textbf{Generative Models} \enskip
Most generative models mainly focus on unimodal generation.
VAE \cite{kingma2013auto}, GAN \cite{goodfellow2014generative}, and GPT \cite{brown2020language} generate image or text without any conditional input.
Recently, these studies have been prominent approaches for text-conditional image \cite{ramesh2021zero, ding2021cogview} and image-conditional text generation \cite{li2020oscar, hu2022scaling}.
With these model variants, unimodal generation has been rapidly improved.
However, studies on multimodal generation are still unexplored.
Joint GAN \cite{shimanuki2019joint} proposes unconditional image-text generation, but this model generates two images, one for the image and another for the text, thus it must undergo the OCR process \cite{smith2007overview}.
Similarly, MMVAE \cite{shi2019variational} generates two images with a shared space on MNIST-SVHN \cite{shi2019variational}. However, both images depict simple digits. They also experiment with an image-text pair dataset. It generates text, but for the image, it retrieves the nearest-neighbor original image in the feature space. In this paper, we use MMVAE with customized decoder for comparison.

\vspace{0.1cm} \noindent
\textbf{Vector Quantized Variational Autoencoder} \enskip
VQ-VAE \cite{oord2017neural} is a representative model that maps continuous input into discrete representations by adopting an encoder-decoder architecture with a fixed size learnable codebook.
In this paper, we aim to jointly generate image-text pairs without any conditional input.
To achieve this, our model requires a continuous or discrete joint representation space, since image is inherently continuous and text is discrete.
While a continuous space is a predominant approach for joint representation learning, we adopt a discrete space for the following reasons.
First, powerful autoregressive generative model \cite{brown2020language} has been developed to model distributions over discrete variables.
Next, the discrete space does not suffer from several drawbacks common to the continuous space, such as posterior collapse in VAE \cite{kingma2013auto}.
In addition, discrete variables have the advantage of being more interpretable and space-efficient than continuous variables \cite{cartuyvels2021discrete}.
Consequently, we propose a simple approach based on VQ-VAE, which learns a quantized joint representation space.


\vspace{0.1cm} \noindent 
\textbf{Joint Image-Text Representations} \enskip
Following the success of Transformer \cite{vaswani2017attention} in NLP tasks, there is a simultaneous explosion of Transformer-based models in joint representation learning.
Previous works (e.g. UNITER \cite{chen2020uniter}, Pixel-BERT \cite{huang2020pixel}, VLP \cite{zhou2020unified}) utilize BERT \cite{devlin2018bert}, a Transformer encoder-based model, to learn joint image-text representations. By leveraging the bidirectional self-attention mechanism of BERT, both images and text can capture the semantic correlation between them without requiring annotations that align image and text.

\section{Multimodal Cross-Quantization VAE (\model)}
\begin{figure*}[h]
    \centering
    \includegraphics[width=\textwidth]{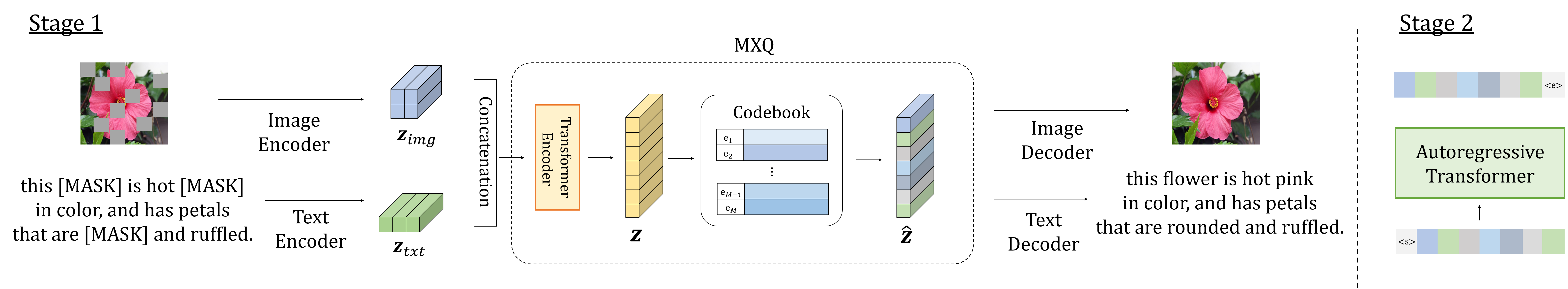}
    \vspace{-7mm}
    \caption{Unconditional Image-Text Pair Generation.
    In Stage 1, \model takes a masked image-text pair as input, and learns a quantized joint representation space. Then, the input is converted into a unified code sequence.
    In Stage 2, Autoregressive Transformer models a joint distribution over the code sequence. At inference, \model decodes a sampled code sequence to an image-text pair.}
    \vspace{-5mm}
\label{fig:model}
\end{figure*}

Our goal is to generate semantically consistent image-text pairs simultaneously without any conditional input.
To achieve this, we learn a joint representation space by quantizing both image and text into a discrete space based on VQ-VAE.
As shown in Fig.~\ref{fig:model}, we adopt a two-stage approach.

\subsection{Stage 1: Learning a Quantized Joint Representation Space}
\label{sec:method_stage1}
\model learns to discretize image-text pair into a quantized joint representation space and reconstructs them.
It consists of three major parts: Encoders, Decoders, and MXQ.
The MXQ module contains the Transformer encoder and a codebook $C = \{e_m\}^M_{m=1}$ of size $M$, where $e_m \in \mathbb{R}^d$.
Each encoder and decoder is 2D CNN for image and 1D CNN for text.
The effect of other architectural choices is discussed in the experiments.\\

\vspace{-2mm}
\noindent
\textbf{Input Masking} \enskip
Given an image-text pair, we first split the image $I \in \mathbb{R}^{H \times W \times 3}$ into non-overlapping patches of equal size and the text into tokens $T=\{t_1, ..., t_N\}$ with WordPiece \cite{wu2016google}. Then, we randomly mask the patches and the tokens with a probability $p$ from a uniform distribution without replacement.
Each pixel in the masked patches is set as zero. We replace the masked tokens with a special token \texttt{[MASK]}.
This approach makes two modalities complementary.
To reconstruct the masked part, the model should not only refer to the unmasked part of each modality (\textit{i.e.}, intra-modal), but also to the unmasked part of the other modality (\textit{i.e.}, cross-modal).
In this way, the model can learn the relationship between image and text.\\

\vspace{-2mm}
\noindent
\textbf{Encoders} \enskip
A masked input image is encoded to a set of image feature map $\mathbf{z}_{img} \in \mathbb{R}^{h \times w \times d}$.
Similarly, a masked text input is encoded to $\mathbf{z}_{txt} \in \mathbb{R}^{n \times d}$, where both of them are downsampled by a factor of $f$; that is, $h = \frac{H}{f}, w = \frac{W}{f}$, $n = \frac{N}{f}$.\\

\vspace{-2mm}
\noindent
\textbf{MXQ} \enskip
The Transformer encoder takes the concatenation of $\mathbf{z}_{img}$ and $\mathbf{z}_{txt}$ as input,
and produces joint image-text representations $\mathbf{z} \in \mathbb{R}^{\ell \times d}$, where $\ell = h \times w + n$.
The output is discretized into the quantized joint space by performing the nearest-neighbor search in the codebook $C$ as given in Eq.~\eqref{eq1} and produces a unified code sequence $\hat{\mathbf{z}} \in \mathbb{R}^{\ell \times d}$.
With this simple approach, a discrete code can contain the correlated information of image and text.
\vspace{-2mm}
\begin{equation}\label{eq1}
     \hat{\mathbf{z}}_i = \text{Quantize}(\mathbf{z}_i) = e_m
    \quad \text{where} \; m = \argmin_j \|\mathbf{z}_i - e_j\|
\end{equation}
\vspace{-4mm}

\noindent
\textbf{Decoders} \enskip
We first apply a linear layer to the spatial dimension (\emph{i.e.}, $\ell$) of $\hat{\mathbf{z}}$ to ensure that the decoder takes the desired size as input and produces $\hat{\mathbf{z}}_{img}$ and $\hat{\mathbf{z}}_{txt}$ for image and text, respectively.
The decoder then reconstructs the original input from $\hat{\mathbf{z}}_{img}$ and $\hat{\mathbf{z}}_{txt}$, yielding reconstruction results, $I' \in \mathbb{R}^{H \times W \times 3}$ and $T'$, respectively.\\

\vspace{-2mm}
\noindent
Our model is optimized using the following objective:
    \begin{equation}
        L = \delta_1\underbrace{\|I - I'\|^2_2}_{\mathclap{\text{image recon loss}}} - \delta_2\underbrace{\log p(T|\hat{\mathbf{z}}_{txt})}_{\mathclap{\text{text recon loss}}} + \delta_3\underbrace{\|sg[\mathbf{z}] - \hat{\mathbf{z}}\|^2_2}_{\text{codebook loss}} + \delta_4\underbrace{\|sg[\hat{\mathbf{z}}] - \mathbf{z}\|^2_2}_{\text{commitment loss}}
    \end{equation}
where each loss term is weighted by $\delta_i$ and $sg$ refers to a stop-gradient.

\subsection{Stage 2: Unconditional Image-Text Pair Generation}
\label{sec:method_stage2}
We adopt the Autoregressive Transformer \cite{radford2019language} architecture to model a joint distribution over the sequence of unified code indices $c = (c_1, c_2, ..., c_\ell)$ from Stage 1.
The probability of each code index in the sequence is conditioned on all previously predicted code indices $c_{<n} = (c_1, c_2, ..., c_{n-1})$ and the joint distribution of the sequence is obtained as the product of conditional distributions: $p(c) = \prod^\ell_{n=1}p(c_n|c_1, c_2, ..., c_{n-1}) =  \prod^\ell_{n=1}p(c_n|c_{<n})$.
During training, \model quantizes the input image-text pair into the unified code sequence, then Autoregressive Transformer is trained to predict the next code index in the given sequence.
At inference, we sample a code sequence from Autogregressive Transformer via Top-\textit{k} sampling \cite{fan2018hierarchical}, then \model decodes the sampled code sequence to an image-text pair.


\section{Experimental Settings}
\subsection{Datasets}
\label{sec:Dataset}
\begin{figure*}[h]
    \centering
    \includegraphics[height=5cm]{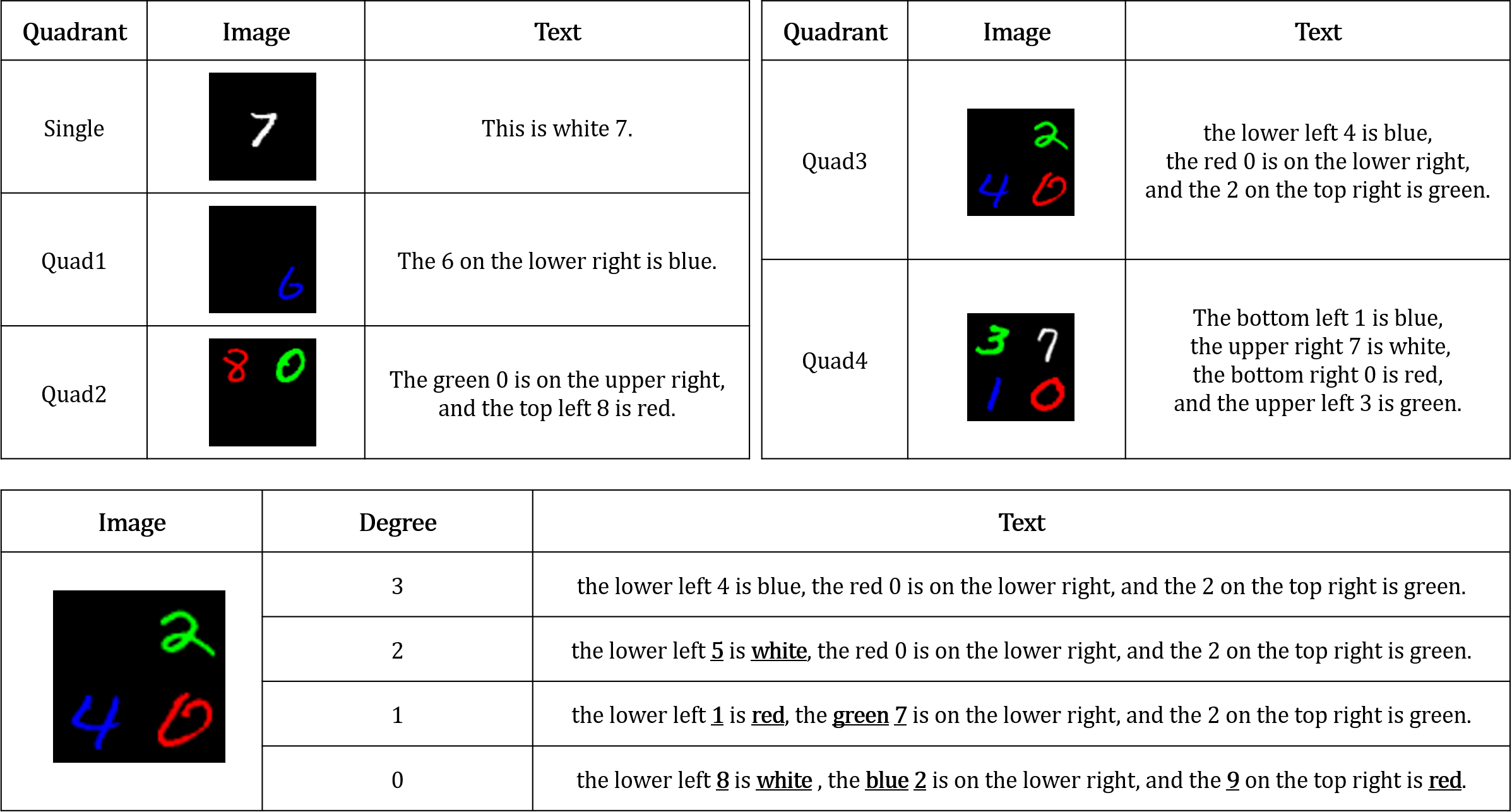}
    \vspace{-3mm}
    \caption{Examples of Caption MNIST (Top) and the Quad3 Degree dataset (Bottom).}
    \vspace{-5mm}
\label{fig:captionmnist}
\end{figure*}

\begin{figure*}[h]
    \centering
    \includegraphics[width=\textwidth]{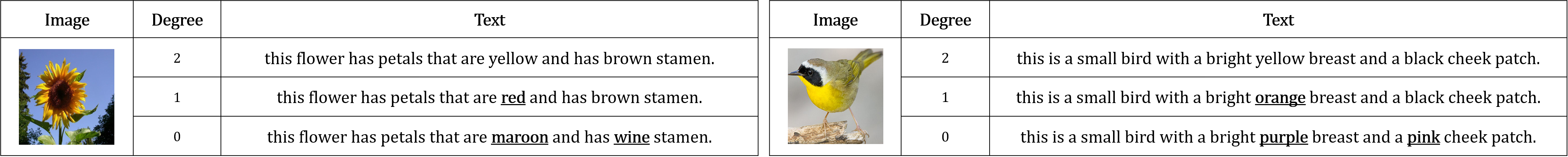}
    \vspace{-7mm}
    \caption{Examples of the Flower (Left) and CUB (Right) Quad2 Degree datasets.}
    \vspace{-5mm}
\label{fig:cub_flower_degree}
\end{figure*}

\textbf{Caption MNIST.} Following \cite{shin2021translation}, we build 600k synthetic image-text pairs. Each pair contains several colors, digits, and positions. We have 4 colors (white, red, green, and blue), 10 digits (0 to 9) and 5 positions (center, top left, top right, bottom left, and bottom right).
According to the filled quadrant, we refer to each subset as Single and Quad1 to Quad4.
For example, Single pairs only have one colored digit at the center of the image and a corresponding caption, whereas Quad3 pairs have colored digits in three quadrants, also with a corresponding caption. See Fig.~\ref{fig:captionmnist} for more details.\\
\textbf{Oxford Flower-102} \cite{nilsback2008automated} (Flower) contains 8,189 flower images with 10 captions per image.\\
\textbf{CUB-200-2011} \cite{wah2011caltech} (CUB) consists of 11,788 bird images with 10 captions per image. We use a bounding box to cut the background of the image and only use the content.\\
\textbf{COCO} \cite{lin2014microsoft} is a real-world dataset with about 120k images and 5 captions per image.\\
\textbf{Degree datasets.} To evaluate the semantic correlation between image and text in the quantized joint space in Stage 1, we construct the Degree dataset by gradually adjusting the degree of alignment between image and text. More specifically, for Caption MNIST, we replace the color and digit in the caption with other random colors and digits. For instance, Quad3 can have 4 degrees from perfectly paired (Degree 3) to completely unpaired (Degree 0). Refer to Fig.~\ref{fig:captionmnist} for more details. For Flower and CUB, variables besides color are difficult to control, thus we only consider the number of unique colors in the caption and replace them with other random colors. According to the number of unique colors in the caption, we refer it as Quad1 to Quad4.
See Fig.~\ref{fig:cub_flower_degree} for more details.

\subsection{Evaluation Metrics}
\subsubsection{Multimodal Semantic Correlation Evaluation}
\label{sec:joint_quantized_space_evaluation}
We evaluate the semantic correlation between image and text in the quantized joint space by the text reconstruction accuracy on the Degree dataset.
Since the quantized joint space captures the semantically correlated parts of images and text, the model should identify the corrupted parts in the Degree dataset and not reconstruct that part \textit{as is}.
For instance, in the Caption MNIST Quad1 Degree dataset, the Degree 1 input text is ``the green 0 is on the upper right.'' and the Degree 0 input text is ``the white 1 is on the upper right.''.
The input image depicts the Degree 1 text in this case.
If the reconstructed text is ``the green 0 is on the upper right.'' for both Degree 1 and 0 input text, the text reconstruction accuracy will be 1 and 0 for each, as we only consider color and digit for calculating text reconstruction accuracy (only color in cases of Flower and CUB).
According to this, the desired accuracy, for instance, would be 1, 0.67, 0.33, and 0 for each degree of the Quad3 Degree dataset.

\subsubsection{Generated Image-Text Alignment Evaluation}
\label{sec:generated_img_txt_alignment_evaluation}
We evaluate the semantic consistency of the generated image-text pairs with rule-based semantic parser on Caption MNIST, label-based modified unigram precision \cite{papineni2002bleu} and sentence similarity on Flower and CUB, and CLIP-based retrieval on COCO.\\
\textbf{Rule-based Semantic Parser.}
Following \cite{shin2021translation}, we extract a set of position, color and digit of the generated text with the rule-based parser. With a color and digit classifier trained on Caption MNIST images that achieved 100\% and 99.5\% accuracy respectively, we predict the color and digit of the position that corresponds to the parsed text in the generated image and measure whether both the predicted color and digit match the parsed text at that position.\\
\textbf{Label-based Evaluations.}
For Flower and CUB, we measure the semantic consistency between the generated caption and all original captions that belong to the same label as the generated image.
Specifically, we first train image classifiers using the original image and achieve 99\% and 93\% accuracy for Flower and CUB, respectively. Then, we predict the label of the generated image and collect all original captions from the same label.
For the modified unigram precision \cite{papineni2002bleu}, we report the average score of the multiset intersection of words in the original text and the generated text divided by the total number of words in the generated text.
For the sentence similarity, a pre-trained BERT \cite{devlin2018bert} takes the generated text and the original text from the same label, separately, and outputs a mean-pooled vector. Then, we calculate the cosine similarity between them and report a Top-1, 5 and 10 averaged score.\\
\textbf{CLIP-based Retreival.}
For COCO, we report the Precision@\{1, 5, 10\} to measure the retrieval accuracy of CLIP \cite{radford2021learning} of the generated text from the 100 text candidates; that is, 1 positive from the generated text, 99 random negatives from the original text.

\subsection{Baselines}
\textbf{\model w/o IM (Input Masking).} The architecture is the same as \model, but without input masking.\\
\textbf{\model w/o TC (Text Compression).} This replaces the 1D convolution-based text encoder and decoder with the Transformer encoder.
Note that the Transformer text encoder outputs the same number of embeddings as the input, contrary to 1D convolutional layers that compresses the input text gradually by each layer. This is why we name it \textit{w/o \textbf{T}ext \textbf{C}ompression}.\\
\textbf{Unimodal Quantizer.} In Stage 1, this baseline discretizes each modality separately. For the image, it follows the original VQ-VAE. Since the text is originally discrete, we directly use the word embeddings. In Stage 2, we concatenate the code sequence of image and text embeddings as the input.
Depending on which modality comes first in Stage 2 input, we refer to it as I\_T$_\text{Embd}$ or T$_\text{Embd}$\_I.\\
\textbf{Only Sharing $C$.} In Stage 1, this baseline only shares the codebook $C$ without the Transformer module that combines image and text together and the input masking. 
In Stage 2, depending on which modality comes first, we refer to it as I\&T or T\&I.\\
The variants of Unimodal Quantizer and Only Sharing $C$ also generate an image-text pair without any conditional input, and the only difference is which modality comes first in Stage 2.\\

\vspace{-5mm}
\subsection{Implementation Details}
\textbf{Stage 1.} For Caption MNIST, the codebook $C$ is $256 \times 128$, the input size is $64 \times 64 \times 3$ image and $64 \times 128$ word embedding. Each input is downsampled by a factor of $f=8$.
We adopt 2 stacks of the Transformer encoder and apply an input masking ratio $p=0.3$.
We use a batch size of 800 with a learning rate of $5 \times 10^{-4}$. We set $\delta_1$, $\delta_2$ and $\delta_3$ to 1.0 and $\delta_4$ to 0.25.\\
\textbf{Stage 2.} We adopt GPT-2 \cite{radford2019language} for an autoregressive generative model with 8 layers, 8 attention heads and 512 embedding dimensions.
We adopt Top-\textit{k} sampling \cite{fan2018hierarchical} with \textit{k} = 10. For Caption MNIST, we use a batch size of 800 with a learning rate of $5e-4$.
In all our experiments, we use AdamW \citep{loshchilov2017decoupled} with $\beta_1=0.9$, $\beta_2=0.99$ with cosine decay learning rate scheduler and train the model using a NVIDIA RTX A6000.
\vspace{-5mm}

\section{Results and Discussion}
\subsection{Multimodal Semantic Correlation Results}
\vspace{-1.5mm}
We first study \textbf{the effectiveness of \model in constructing multimodal semantic correlation in the quantized joint space} as described in Sec.~\ref{sec:joint_quantized_space_evaluation}.
Tab.~\ref{tab:stage1_all} shows the results.
On the Caption MNIST Degree dataset, we observe that Only Sharing $C$ and \model w/o TC cannot fully capture the semantic correlation between image and text.
In fact, \model w/o TC completely fails to learn the relationship between image and text.
Our approach, on the other hand, shows the best approximation in all quadrants.
This result suggests that \model can identify the correlation and the difference between image and text.
Also, we see the advantage of the input masking that brings in considerable improvement.
Moreover, our approach again achieves the best performance over baselines on the Flower and CUB Degree datasets.
See more results in supplementary.
The above results show that the Transformer encoder for cross-modal attention and the text compression are essential for the multimodal semantic correlation, and the input masking also plays a significant role.
Consequently, we choose \model as our final design for unconditional image-text pair generation in Stage 2.

\begin{figure*}[!htb]
    \centering
    \includegraphics[height=6cm]{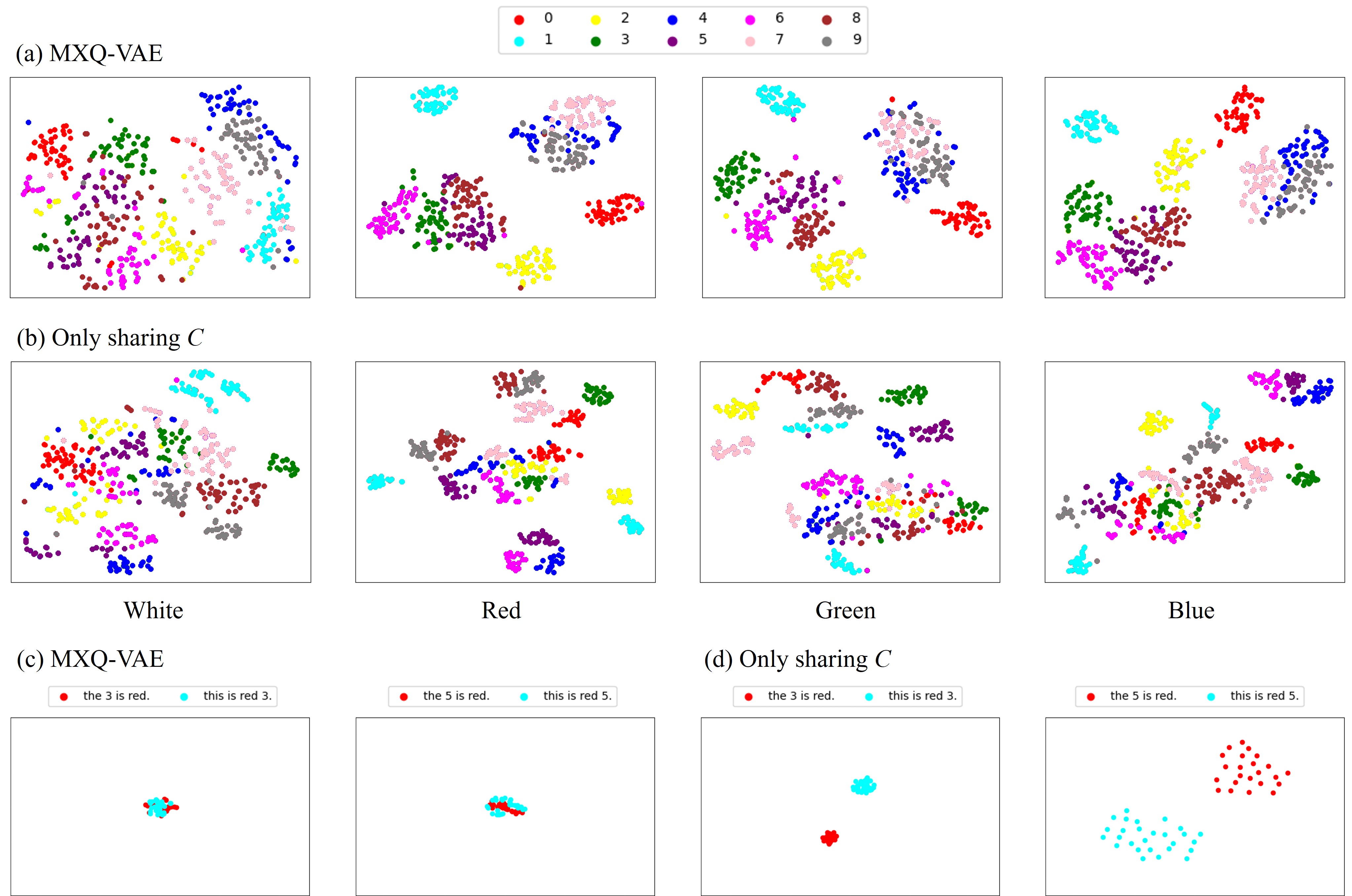}
    \caption{t-SNE visualizations of the unified code sequence on Caption MNIST Single image-text pairs. In (a) and (b), \model has a unique cluster for each digit compared to the baseline. In (c) and (d), unlike \model, the baseline cannot identify the correlation between the two types of text, even though they contain the same content.}
    \vspace{-5mm}
\label{fig:latent_viz}
\end{figure*}

\begin{figure*}[!htb]
    \centering
    \includegraphics[width=\textwidth]{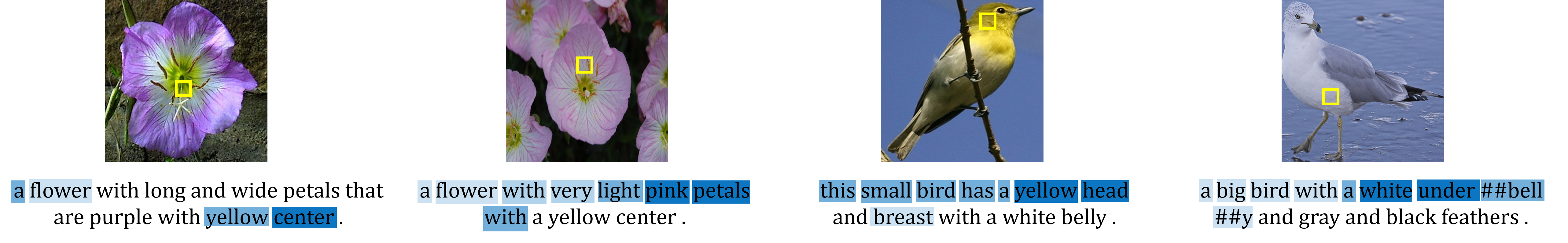}
    \vspace{-7mm}
    \caption{Visualization of the attention maps on Flower and CUB. The deeper color indicates a higher attention score. Although we do not provide any explicit guidance, \model can properly attend to the related tokens that correspond to the image patch.}
    \vspace{-5mm}
\label{fig:attn_viz}
\end{figure*}

We also visualize the unified code sequences and the attention maps of the Transformer encoder in the MXQ module.
Fig.~\ref{fig:latent_viz} (a) and (b) show the t-SNE \cite{van2008visualizing} visualizations of the unified code sequences with \model and Only sharing $C$ for 10 digits per color on Caption MNIST Single image-text pairs.
\model has a unique cluster for each digit, while the baseline has two.
This is because there are two types of text in the Single pairs: 1) This \{digit\} is \{color\}; 2) This is \{color\} \{digit\}.
As shown in Fig.~\ref{fig:latent_viz} (c) and (d), \model can capture the correlation between them, but the baseline completely fails even though they contain the same content.
Fig.~\ref{fig:attn_viz} visualizes the attention maps on the text tokens when the image patch is given as a query.
These results again show the superiority of \model.
\vspace{-1mm}
\begin{table}[hbt!]
\begin{adjustbox}{height=4cm,center}
\begin{tabular}{ccccccccc}
\toprule
    \multicolumn{1}{c}{Dataset} &
    \multicolumn{1}{c}{Degree} &
    \multicolumn{1}{c}{Models}  &
    \multicolumn{1}{c}{Degree 4} &
    \multicolumn{1}{c}{Degree 3} &
    \multicolumn{1}{c}{Degree 2} &
    \multicolumn{1}{c}{Degree 1} &
    \multicolumn{1}{c}{Degree 0} \\
    
    \midrule
    & & Only Sharing \emph{$C$}     & 0.486 & 0.443 & 0.394 & 0.358 & 0.315 \\
    & & \model w/o TC   &	1.0	&	0.975	&	0.951	&	0.929	&	0.906 \\
    & Quad4 & \model w/o IM                 &	0.896	&	0.802	&	0.698	&	0.595	&	0.498 \\
    Caption MNIST& &  \textbf{\model (Ours)}                         &	 0.969	&	 0.729	&	 0.489	&	 0.248	&	 0.009 \\
    
    \cmidrule(lr){2-8}
    & & Only Sharing \emph{$C$}     & - & 0.713 & 0.64 & 0.558 & 0.483 \\
    & & \model w/o TC   & - &	1.0	&	0.968	&	0.943	&	0.909 \\
    & Quad3 & \model w/o IM                  & - &	0.999	&	0.862	&	0.714	&	0.564 \\
    & &  \textbf{\model (Ours)}                          &	-	&	 0.997	&	 0.663	&	 0.334	&	 0.012 \\
    
    \midrule
    & & Only Sharing \emph{$C$}     &	0.939	&	0.704	&	0.516	&	0.321	&	0.131 \\
    & & \model w/o TC &	1.0	&	0.944	&	0.886	&	0.866 & 0.810 \\
    & Quad4 & \model w/o IM                  &	0.997	&	0.728	&	0.482	&	0.278	&	0.067 \\
    Flower & & \textbf{\model (Ours)}                          &	0.996	&	0.737	&	0.490	&	0.250	&	0.014 \\
    
    \cmidrule(lr){2-8}
    & & Only Sharing \emph{$C$}     & - &	0.959	&	0.675	&	0.426	&	0.158 \\
    & & \model w/o TC & - &	1.0	&	0.927	&	0.870	&	0.816 \\
    & Quad3 & \model w/o IM                  & - &	0.997	&	0.662	&	0.377	&	0.090 \\
    & & \textbf{\model (Ours)}                          & - &	0.999	&	0.660	&	0.339	&	0.019 \\
    
    \midrule
    & & Only Sharing \emph{$C$}     &	0.985	&	0.771	&	0.572	&	0.356	&	0.155 \\
    & & \model w/o TC &	1.0	&	0.948	&	0.894	&	0.825	&	0.748 \\
    & Quad4 & \model w/o IM                  &	0.998	&	0.833	&	0.645	&	0.424	&	0.181 \\
    CUB & & \textbf{\model (Ours)}                          &	0.995	&	0.749	&	0.515	&	0.292	&	0.083 \\
    
    \cmidrule(lr){2-8}
    & & Only Sharing \emph{$C$}     & - &	0.986	&	0.755	&	0.498	&	0.199 \\
    & & \model w/o TC & - &	1.0	&	0.938	&	0.864	&	0.772 \\
    & Quad3 & \model w/o IM                  & - &	0.998	&	0.806	&	0.559	&	0.248 \\
    & & \textbf{\model (Ours)}                          & - &	0.995	&	0.709	&	0.421	&	0.119 \\
    
    
\bottomrule
\end{tabular}
\end{adjustbox}
\caption{Multimodal semantic correlation on the Caption MNIST, Flower and CUB Degree datasets. The scores close to 1.0, 0.75, 0.5, 0.25, 0.0 on Quad4 and 1.0, 0.67, 0.33, 0.0 on Quad3 are better. In general, our approach shows the best performance.}
\label{tab:stage1_all}
\end{table}




\subsection{Generated Image-Text Alignment Results}
We evaluate \textbf{the semantic consistency of the generated image-text pairs} as described in Sec.~\ref{sec:generated_img_txt_alignment_evaluation}.
Tab.~\ref{tab:mnist_gen_alignment} shows the results on Caption MNIST. We can observe that \model outperforms all baselines on every quadrant.
Also, note that all baselines are vulnerable to which modality is given first to generate the image-text pairs.
All models given text first significantly underperform up to 21.7\% (in I\_T$_\text{Embd}$ and T$_\text{Embd}$\_I) on average compared to the model given image first.
We assume that this is due to the fact that quantized image is longer than text and image often contains more complex (and complete) information than text.
\model, on the other hand, avoids this problem with a unified code sequence.
We report the results on Flower and CUB in Tab.~\ref{tab:cub_flower_stage2}.
Compared to the baselines, \model performs well in all metrics, indicating its capability to generate semantically consistent image-text pairs for real-world data as well as carefully controlled synthetic data.
Also, we again demonstrate the superiority of \model on COCO in Tab.~\ref{tab:coco_gen_alignment}.
This result indicates that our approach can be extended to large-scale data in the future.

\begin{figure*}[!htb]
    \centering
    \includegraphics[width=\textwidth]{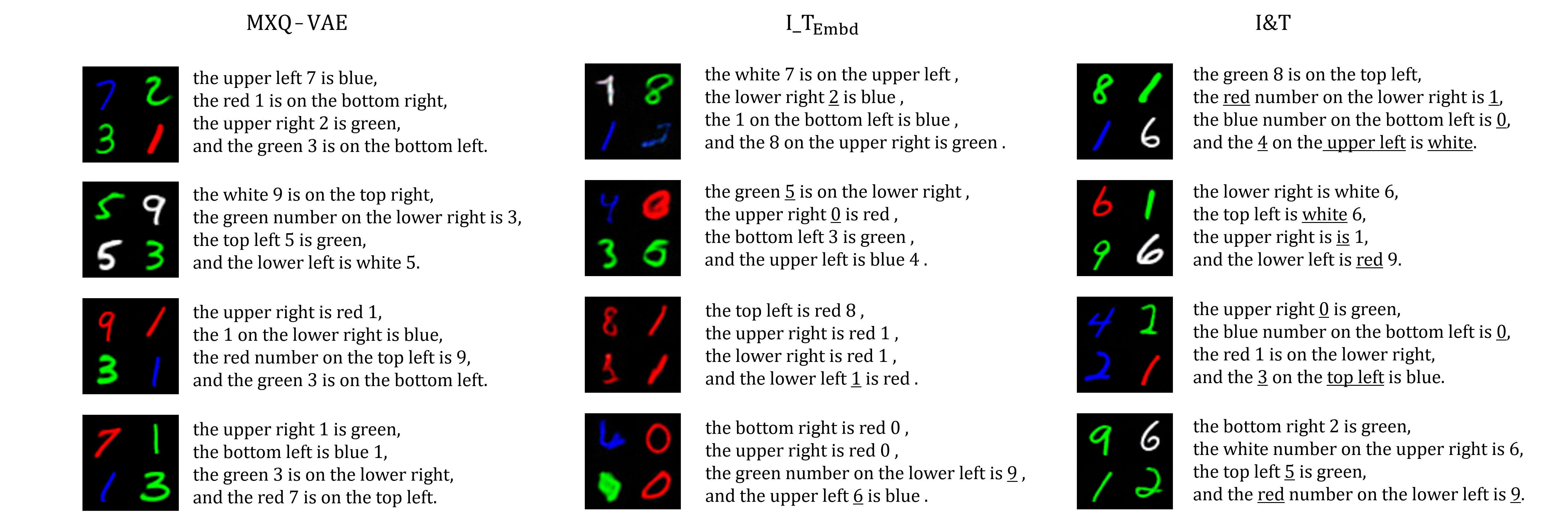}
    \caption{Comparison of generated image-text pairs on Caption MNIST}
\label{fig:gen_mnist_comparison}
\end{figure*}

\begin{figure*}[!htb]
    \centering
    \includegraphics[height=7cm]{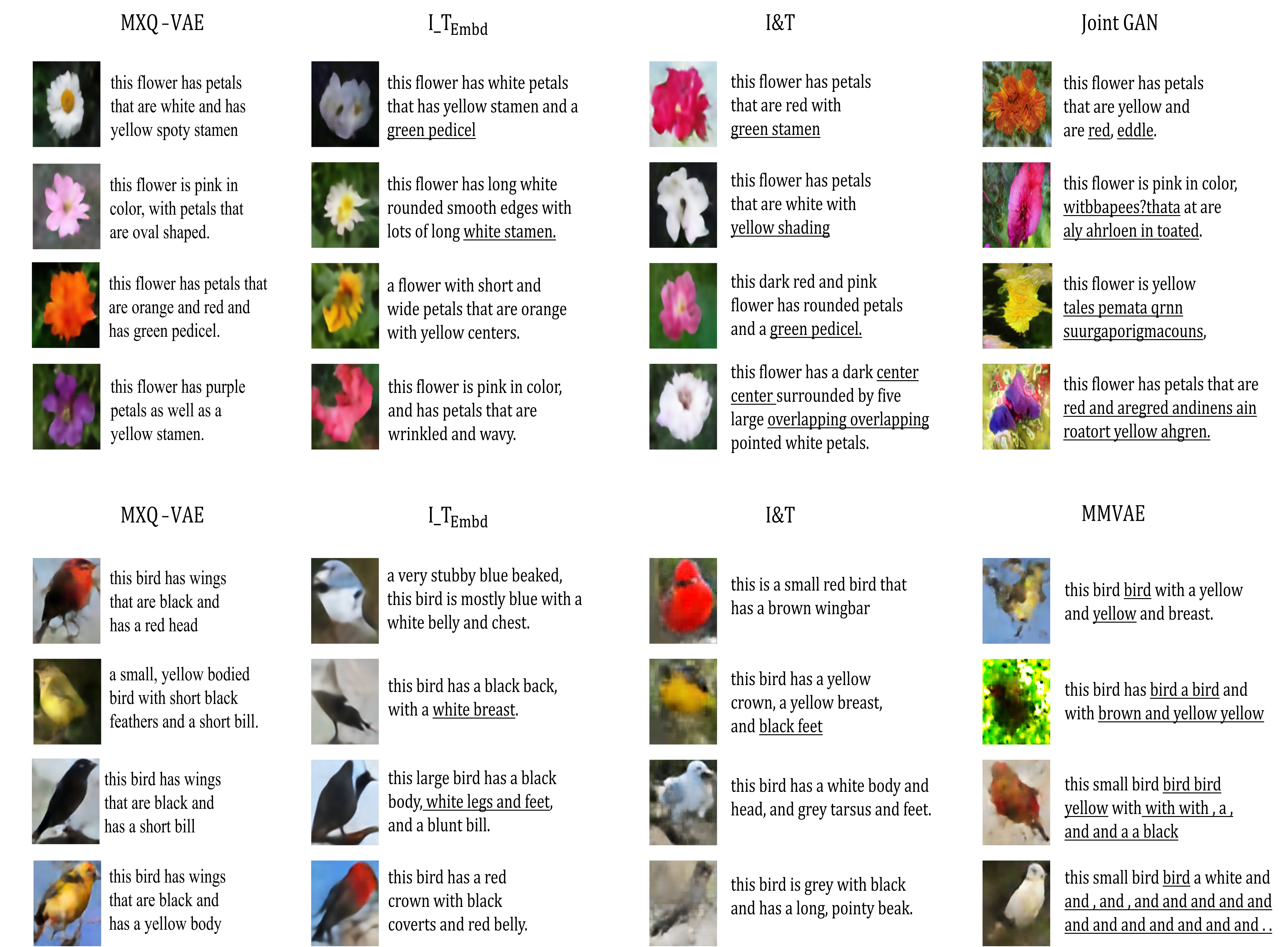}
    \caption{Comparison of generated image-text pairs on Flower and CUB}
\label{fig:gen_flower_cub_comparison}
\end{figure*}

Fig.~\ref{fig:gen_mnist_comparison} shows the comparison of generated image-text pairs on Caption MNIST. We can see that \model can generate semantically consistent image-text pairs and also show high-fidelity image and text.
In contrast, although I\_T$_\text{Embd}$ can generate high-quality text, the generated images are partially blurry and hard to identify the digit.
Also, I\&T can generate high fidelity images, but the generated pairs are irrelevant, showing its poor ability to capture the relationship between image and text.
Similarly, \model shows its superiority on Flower and CUB as shown in Fig.~\ref{fig:gen_flower_cub_comparison}.
Unlike \model, some parts of the images of I\_T$_\text{Embd}$ are incomplete.
Also, I\&T and MMVAE \citep{shi2019variational} cannot generate detailed parts, and most of the generated images of bird are blurry, and they generate overlapping words.
In addition, Joint GAN \citep{shimanuki2019joint} fails to generate realistic images and the generated text is grammatically incorrect.
The underline in the figure represents the mismatched part between image and text. For more samples, refer to supplementary.
\begin{table}[hbt!]
\begin{adjustbox}{height=1.2cm,center} 
\begin{tabular}{ccccccc}
\toprule
    \multicolumn{1}{c}{Models} &
    \multicolumn{1}{c}{Single} &
    \multicolumn{1}{c}{Quad1} &
    \multicolumn{1}{c}{Quad2} &
    \multicolumn{1}{c}{Quad3} &
    \multicolumn{1}{c}{Quad4} &
    \multicolumn{1}{c}{Average} \\
    \midrule
    I\&T                  & 0.979 & 0.926 & 0.675 & 0.434 & 0.255 & 0.654\\
    T\&I                  & 0.803 & 0.780  & 0.458 & 0.282 & 0.161 & 0.497\\
    I\_T$_\text{Embd}$         & 0.953 & 0.953 & 0.956 & 0.958 & 0.849 & 0.945\\
    T$_\text{Embd}$\_I          & 0.086 & 0.895 & 0.913 & 0.916 & 0.828 & 0.728\\
    \textbf{\model (Ours)}                 &	\textbf{0.998}	&	\textbf{0.997}	&	\textbf{0.994}	&	\textbf{0.996}	&	\textbf{0.974}	&	\textbf{0.992}\\
\bottomrule
\end{tabular}
\end{adjustbox}
\caption{Semantic consistency of the generated image-text pairs on Caption MNIST}
\label{tab:mnist_gen_alignment}
\end{table}

\begin{table}
\vspace{-2mm}
\begin{adjustbox}{width=\textwidth,center}
\begin{tabular}{ccccccccc}
\toprule
    \multirow{3}{*}{Models}  &
    \multicolumn{4}{c}{Flower}  &
    \multicolumn{4}{c}{CUB}  \\
    \cmidrule(lr){2-5}
    \cmidrule(lr){6-9}
    & Modified unigram &  & Sentence similarity & & Modified unigram & & Sentence similarity & \\ 
    \cmidrule(lr){3-5}
    \cmidrule(lr){7-9}
    & precision & Top-1 & Top-5 & Top-10 & precision & Top-1 & Top-5 & Top-10 \\ 
    \midrule
    I\&T                                       &	0.420	&	0.935	&	0.920	&	0.910 & 0.418	&	0.912	&	0.889	&	0.875 \\
    T\&I                                        &	0.402	&	0.910	&	0.891	&	0.880 &	0.425	&	0.917	&	0.896	&	0.884 \\
    I\_T$_\text{Embd}$                                 & 0.406 &	0.938	&	0.923	&	0.913 &	0.425	&	0.921	&	0.900	&	0.886 \\
    T$_\text{Embd}$\_I                                 &	0.392	&	0.927	&	0.900	&	0.884 &	0.423	&	0.917	&	0.895	&	0.882 \\
    Joint GAN \citep{shimanuki2019joint} $^\ast$    & 0.324 & 0.808	&	0.788	&	0.774 & - & - & - & - \\
    MMVAE \cite{shi2019variational} & - & - & - & - & 0.262 & 0.707 & 0.682 & 0.667 \\
    \textbf{\model (Ours)} & \textbf{0.428}	& \textbf{0.941}	& \textbf{0.926}	& \textbf{0.916} & \textbf{0.478} & \textbf{0.948} & \textbf{0.919}	& \textbf{0.900} \\
    
\bottomrule
\end{tabular}
\end{adjustbox}
\caption{Semantic consistency of the generated image-text pairs on Flower and CUB.
\\\centering * means that we measure the score with the image-text pairs reported in Joint GAN \citep{shimanuki2019joint}.}
\vspace{-2mm}
\label{tab:cub_flower_stage2}
\end{table}
\begin{table}[hbt!]
\begin{adjustbox}{height=0.7cm,center}
\begin{tabular}{cccc}
\toprule
    \multicolumn{1}{c}{Models} &
    \multicolumn{1}{c}{P@1} &
    \multicolumn{1}{c}{P@5} &
    \multicolumn{1}{c}{P@10}\\
    \midrule
    I\&T               & 0.062 & 0.263 & 0.328 \\
    I\_T$_\text{Embd}$               & 0.083 & 0.29 & 0.425 \\
    \textbf{\model (Ours)}      & \textbf{0.106} & \textbf{0.323} & \textbf{0.491} \\
\bottomrule
\end{tabular}
\end{adjustbox}
\caption{Semantic consistency of the generated image-text pairs on COCO}
\label{tab:coco_gen_alignment}
\end{table}


In summary, extensive results demonstrate that the quantized joint space with multimodal semantic correlation is effective for semantically consistent image-text pair generation.
Furthermore, our approach enhances the meaningful semantic correlation between image and text and also encourages the generation of semantically consistent image-text pairs.


\section{Conclusion and Future Work}
In this study, we propose for the first time a novel vector quantizer, \model, that learns a quantized joint representations space for unconditional image-text pair generation.
We demonstrate that the quantized joint space with a multimodal semantic correlation encourages the generation of semantically consistent image-text pairs.
With extensive experiments, our approach shows superiority in these two aspects over several baselines.
We hope our work suggests new directions for multimodal generation and joint representation learning.
Besides exploring AI's multimodal creativity, one promising application of our approach is training conditional generative models with three or more modalities (\textit{e.g.}, image, text, audio).
By considering \model as a multimodal tokenizer, generative models can generate other modalities given a multimodal code sequence (or vice versa), which remains as future work.

\section*{Acknowledgements}
This work was supported by the KAIST-NAVER Hyper-Creative AI Center
and the Institute of Information \& Communications Technology Planning \& Evaluation (IITP) grants (No.2019-0-00075 Artificial Intelligence Graduate School Program(KAIST) and No.2021-0-01778 Development of human image synthesis and discrimination technology below the perceptual threshold),
and National Research Foundation of Korea (NRF) grant (NRF-2020H1D3A2A03100945 and NRF-2021H1D3A2A03038607)
funded by the Korea government (MSIT).
\bibliography{egbib}

\end{document}